\documentclass[runningheads]{llncs}
\usepackage[T1]{fontenc}
\usepackage{hyperref}
\usepackage{orcidlink}
\usepackage{marvosym}
\usepackage{amsmath}
\usepackage{amssymb}
\usepackage{booktabs}
\usepackage{graphicx}
\begin{document}
\title{NeurGO: Learning to Generate Elite Candidates for Meta-Black-Box Expensive Optimization}
\titlerunning{NeurGO: Generating Elite Candidates for Meta-Black-Box Optimization}
%
\author{Jintao He\orcidlink{0000-0003-1298-7458} \and
Huixiang Zhen\orcidlink{0000-0001-7866-6580} \and
Wenyin Gong\,\textsuperscript{\href{mailto:wygong@cug.edu.cn}{(\Letter})}\,\orcidlink{0000-0003-1610-6865}}
\authorrunning{J. He et al.}
%
\institute{China University of Geosciences, Wuhan, China \\
\email{wygong@cug.edu.cn}
}
\maketitle              
\begin{abstract}
Expensive black-box optimization is ubiquitous in science and engineering, where function evaluations are costly and the evaluation budget is limited.
Traditional evolutionary algorithms and Meta-Black-Box Optimization (MetaBBO) approaches typically consume most evaluations on candidate selection, often wasting precious budget on inferior solutions. 
Although surrogate-assisted evolution and Bayesian optimization aim to reduce evaluations through surrogate models, constructing an accurate global model from limited data remains challenging, and model bias can easily trap the search in local optima.
To overcome these limitations, we propose NeurGO, a generative MetaBBO framework that directly synthesizes elite candidates from historical population states.
Specifically, we employ an attention-based encoder to capture the population-level search trend and condition a decoder on this representation to generate high-quality candidates, avoiding the expensive evaluation of large offspring pools.
We then design a quality-diversity loss to maintain solution quality and population diversity throughout the search.
Through extensive benchmarking on CEC 2008 and the COCO BBOB test suites, our method achieves better optimization performance under the same evaluation budget and exhibits faster convergence.
\end{abstract}
\section{Introduction}

Black-box optimization (BBO) considers global optimization problems where the objective is accessible only through function evaluations and gradient information is unavailable~\cite{Alarie21}. Such problems arise in applications such as hyperparameter tuning~\cite{Wang19,Akiba19} and engineering design~\cite{Pierret99}, where evaluations can be time-consuming or computationally expensive. Under a strict evaluation budget, the central challenge is to approach the global optimum with as few evaluations as possible. Although meta-learning-based optimization algorithms~\cite{Zhang23} have shown strong potential in learning transferable search strategies from historical data, their application to expensive optimization remains relatively underexplored~\cite{Li22}.

\begin{figure}[t]
    \centering
    \includegraphics[width=0.72\linewidth]{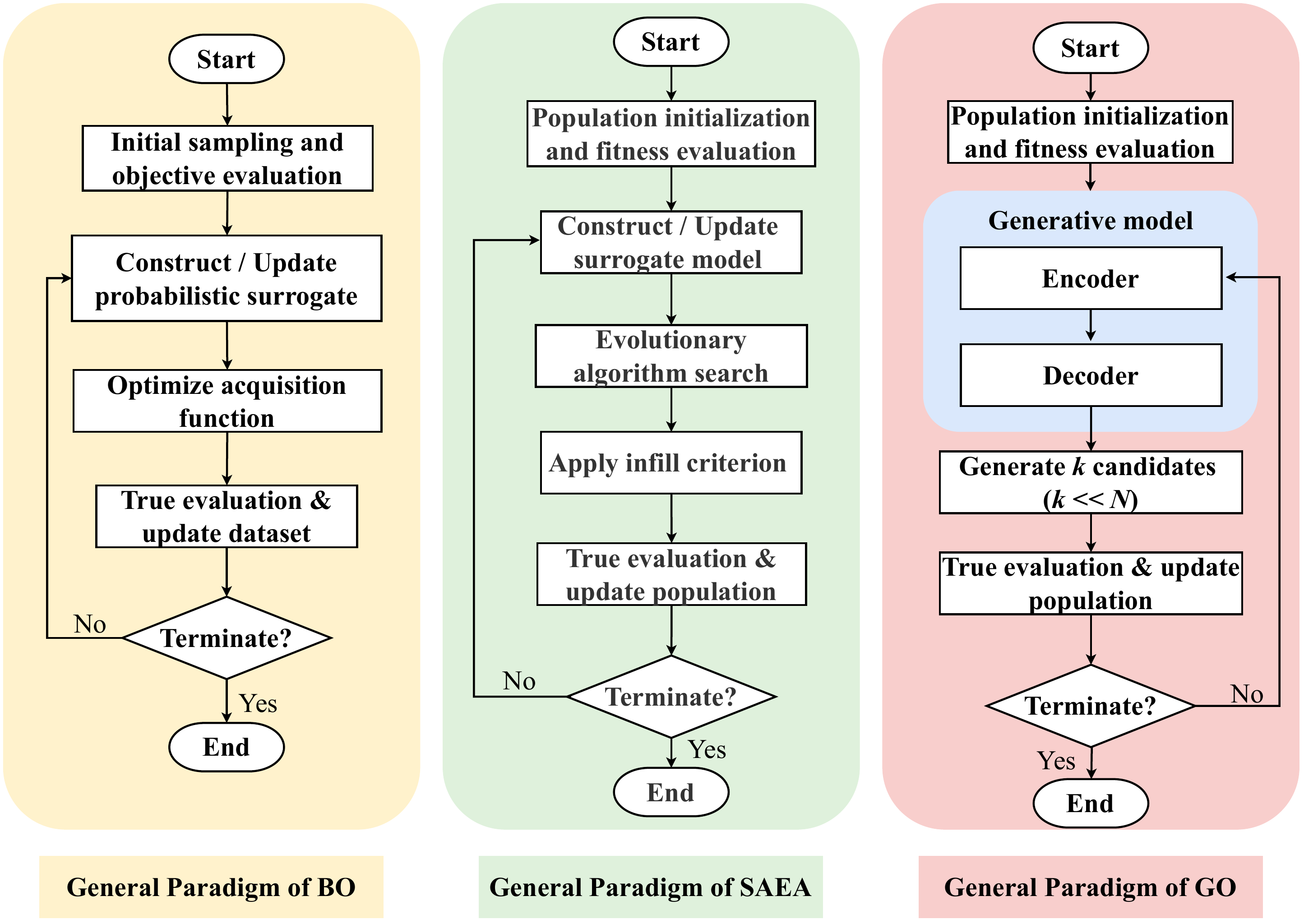}
    \caption{Comparison of three optimization paradigms for expensive black-box optimization. \textbf{Left:} Bayesian optimization. \textbf{Middle:} Surrogate-assisted evolutionary algorithm. \textbf{Right:} Generative optimization.}
    \label{fig:paradigm_comparison}
\end{figure}

As summarized in Figure~\ref{fig:paradigm_comparison}, existing methods for expensive BBO mainly follow two paradigms: \textbf{Bayesian Optimization (BO)}~\cite{Shahriari16} and \textbf{Surrogate-Assisted Evolutionary Algorithms (SAEAs)}~\cite{He23}. Both reduce evaluation cost by relying on surrogate models, but their performance depends heavily on surrogate fidelity~\cite{Kandasamy16}. Under tight budgets, limited observations can lead to inaccurate surrogate guidance, especially in high-dimensional, multimodal, or noisy settings~\cite{Zheng22}. In addition, many such frameworks still rely on hand-crafted heuristics and operators~\cite{Jariego21}, which restricts their adaptability across tasks and budgets. These limitations motivate an alternative direction: learning transferable search priors through cross-task training rather than relying on accurate online surrogates.

Meta-Black-Box Optimization (MetaBBO) has emerged as a promising paradigm for learning generalizable optimization strategies from historical tasks~\cite{Ma25}. However, many existing MetaBBO methods still treat evolutionary algorithms (EA) as the underlying optimizer~\cite{Ma23,Shem-Tov24}, and therefore spend a substantial portion of the evaluation budget on generating and evaluating offspring. In contrast, the generative optimization (GO) paradigm (Figure~\ref{fig:paradigm_comparison}, right) directly produces only a few high-value candidates for expensive evaluation, making it particularly suitable for low-budget expensive optimization.

To address these challenges, we propose \textbf{NeurGO}, a generative meta-learning framework for expensive black-box optimization that directly synthesizes a small set of high-quality candidates from the current population. NeurGO consists of a \textbf{Population Context Encoder (PCE)} and an \textbf{Elite Synthesis Decoder (ESD)}. PCE captures population-level search information from the current population, while ESD maps the resulting representation to a small set of candidate solutions. During meta-training, we optimize a \textbf{Quality--Diversity (QD)} objective that promotes both candidate quality and decision-space diversity. We provide the source code of NeurGO online\footnote{Source code available at: \url{https://github.com/spider-queen/NeurGO}}.

The main contributions of this paper are summarized as follows:
\begin{itemize}
    \item \textbf{Generative MetaBBO for expensive optimization.} We propose NeurGO, a generative MetaBBO framework that improves evaluation efficiency by directly synthesizing a few high-value candidates under strict evaluation budgets.
    \item \textbf{End-to-end architecture with QD objective.} We introduce an end-to-end generative framework that learns a mapping from population states to candidate solutions while balancing quality and diversity.
    \item \textbf{Extensive empirical evaluation.} We compare NeurGO against EA, SAEA, BO, and MetaBBO baselines, and show that it achieves faster convergence and better final performance under tight evaluation budgets.
\end{itemize}

\section{Related work}
\subsection{Surrogate-Assisted Expensive Optimization}
SAEA reduces the number of expensive objective evaluations by integrating inexpensive surrogate models into the iterative search~\cite{Li23a}. 
Existing SAEA are commonly grouped into single-surrogate and multi-surrogate variants~\cite{Liang25}, depending on the number of surrogates and how they are coordinated. 
Single-surrogate methods typically construct a global model to approximate the fitness landscape over the search space~\cite{Gu24}. 
BO is a canonical example: it uses Gaussian processes to build a probabilistic surrogate and selects evaluations by optimizing an acquisition function that trades off exploration and exploitation~\cite{Wang23}. 
Multi-surrogate methods, in contrast, combine several surrogates of the same or different types to guide the search, aiming for more robust predictions and better-calibrated uncertainty estimates than a single model~\cite{Hu22}.

Despite substantially reducing evaluation cost, SAEAs have notable limitations. First, their performance depends strongly on surrogate accuracy for the target task. 
In the early stages of optimization, limited data often induces systematic prediction bias, a challenge commonly known as the cold-start problem~\cite{Xue25,Lyu24}. Second, most SAEAs train surrogates from scratch for each new task, which limits cross-task generalization~\cite{Liu24}. As a result, they cannot effectively leverage priors learned from historical tasks to steer the search, making it hard to identify promising regions under a severely limited evaluation budget.

\subsection{Meta-Black-Box Expensive Optimization}
MetaBBO has recently emerged as a data-driven paradigm that aims to automatically learn generalizable optimization strategies from historical tasks using machine learning, especially reinforcement learning, thereby reducing reliance on hand-crafted priors~\cite{Yang25}.
Unlike conventional methods that restart the search from scratch for each new task, MetaBBO leverages meta-learning to extract transferable search knowledge from a broad distribution of past tasks so as to accelerate convergence on target tasks~\cite{Ma25a,Song24}.
Early studies mainly employed recurrent neural networks (RNN)~\cite{Chen17,Wichrowska17} to learn parameter update rules analogous to gradient descent, while subsequent works introduced reinforcement learning (RL) to enable dynamic adjustment for algorithm configuration~\cite{Xue22,Adriaensen22}.
Although these methods can generalize across previously seen tasks, relatively few explicitly target expensive optimization.
We next review representative efforts that combine MetaBBO with expensive optimization.

Recent work such as B2Opt~\cite{Li25} introduces Transformer architectures to model operators in genetic algorithms (GA)~\cite{Holland92}, improving performance under tight evaluation budgets.
Moreover, in expensive multi-objective optimization, DB-SAEA~\cite{Du26} incorporates meta-learning to characterize the landscape in both the decision and objective spaces and dynamically coordinates surrogate models via a dual-control mechanism to balance convergence and diversity.
However, even with Transformer or meta-learning enhancements at the meta-level, these methods still follow the bi-level closed-loop structure of MetaBBO ~\cite{Lange23}.
Under tight evaluation budgets, the meta-level policy relies on sparse and noisy feedback from real evaluations ~\cite{Ma25}, which can delay or misguide its decisions. Consequently, the low-level search often requires multiple population iterations, preventing stable performance gains from limited evaluations.
Therefore, a generative framework that directly synthesizes elite candidates from data is crucial to overcome this efficiency bottleneck.

\section{Preliminaries}
This section briefly defines the expensive black-box optimization problem.
We consider the problem of minimizing a black-box objective function: $\Omega \to \mathbb{R}$ over a continuous search space $\Omega \subseteq \mathbb{R}^D$. 
The expression for the objective function $f$ is unknown and its gradient is inaccessible. 
In the low-budget regime, the optimization process is subject to a strict evaluation budget $B_{\max}$.
Here $x_i\in\Omega$ denotes the $i$-th evaluated point and $y_i=f(x_i)$ is its corresponding objective value. Under a strict evaluation budget $B_{\max}$, we seek:
\begin{equation}
    x^{\mathrm{opt}} = \mathop{\arg\min}_{x \in \Omega} f(x), \quad \text{s.t.} \quad |\mathcal{H}_t| \leq B_{\max}
\end{equation}
where \(|\mathcal{H}_t|\) denotes the number of evaluated points. Accordingly, our algorithm is designed to attain high-quality solutions within a limited evaluation budget.

\section{Methodology}
\subsection{Overview of NeurGO}
To address the limited evaluation budget in expensive black-box optimization, we propose NeurGO, an end-to-end generative MetaBBO framework built on a Transformer backbone~\cite{Vaswani17}. Given the current population and their fitness values as input, NeurGO generates a batch of high-quality candidate solutions for costly evaluation.

\begin{figure}[htbp]
    \centering
    \includegraphics[width=\linewidth]{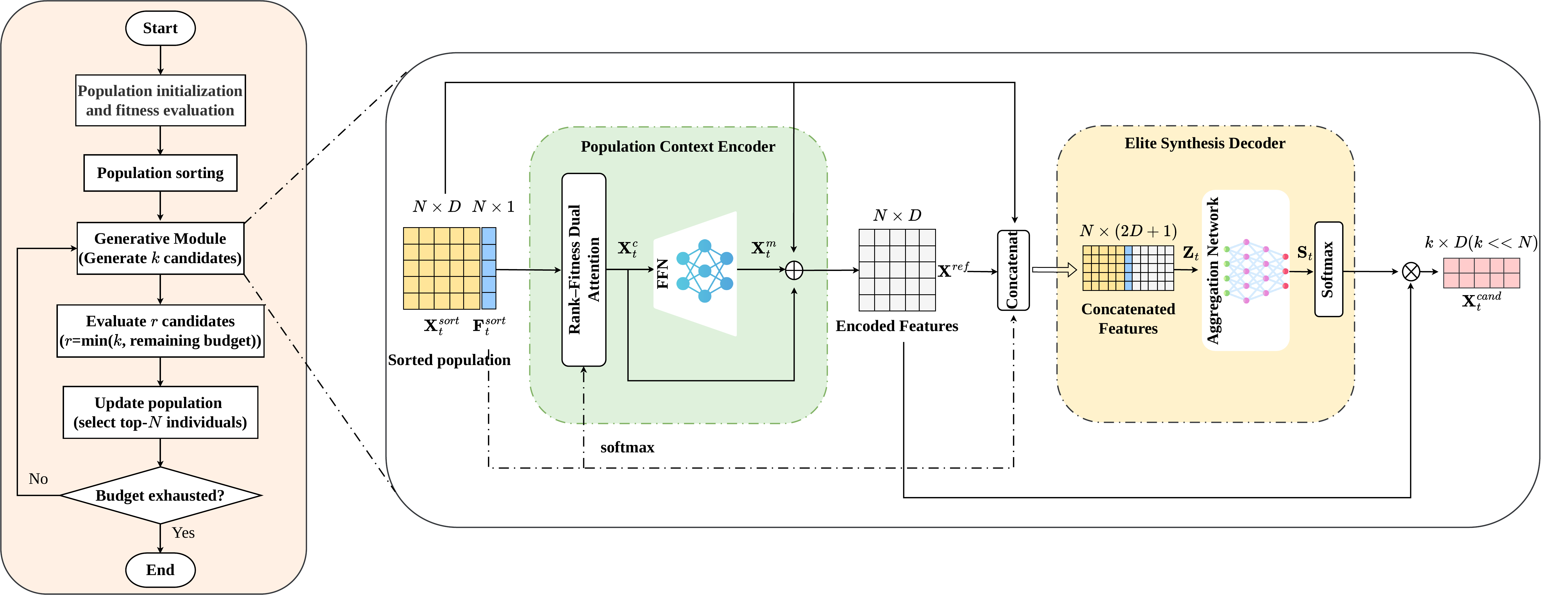}
    \caption{Workflow of NeurGO. Conditioned on the current population and fitness values, the \textbf{PCE} encodes population-level context via self-attention, and the \textbf{ESD} fuses multi-source features to synthesize a batch of elite candidates. The candidates are then expensively evaluated and injected back into the population under the budget constraint. NeurGO is meta-trained with a \textbf{QD} loss to promote both high-quality and diverse proposals.}
    \label{fig:neurgo_framework}
\end{figure}

NeurGO consists of two core modules, \textbf{PCE} and \textbf{ESD}. We organize the overall procedure into three stages: encoding, decoding, and meta-training. In the encoding stage, PCE leverages self-attention to capture global interactions among individuals within the population as well as fitness landscape characteristics and maps them into high-dimensional feature representations that encode evolutionary trends. 
In the decoding stage, ESD adopts a multi-source feature fusion scheme by concatenating the original population, fitness values, and the encoded feature vectors. 
It then computes synthesis weights via a deep multi-layer perceptron to construct a set of high-quality elite candidates. In the meta-training stage, we introduce a \textbf{QD loss} to balance convergence speed and search diversity by maximizing both the quality of generated solutions and their distributional discrepancy. Figure~\ref{fig:neurgo_framework} provides an overview of this pipeline, highlighting how PCE encodes population-level context and how ESD synthesizes elite candidates that are evaluated and injected back into the evolutionary loop under the budget constraint.

\subsection{Population Context Encoder}

The PCE serves as the perception module of NeurGO.
It extracts a global representation of the current search state from population-level observations, including candidate locations and their corresponding fitness values.
By modeling the population as a sequence, PCE leverages the Transformer’s contextual modeling capacity to produce features that capture global interactions among candidates and evolutionary trends.

\textbf{Population sorting.}
We sort the population by fitness before feeding it into the model. This preprocessing step serves as an implicit relative positional encoding, allowing the attention mechanism to prioritize higher-performing individuals. 
Given the population at iteration \(t\), we denote the population matrix by \(\mathbf{X}_t=[x_{t,1},\ldots,x_{t,N}]^\top \in \mathbb{R}^{N\times D}\) and the corresponding fitness vector by \(\mathbf{F}_t=[f(x_{t,1}),\ldots,f(x_{t,N})]^\top \in \mathbb{R}^{N \times 1}\).
We sort the population \(\mathbf{X}_t\) in ascending order of fitness and let \(\mathbf{F}_t^{\mathrm{sort}}\) be the corresponding fitness vector of the sorted population \(\mathbf{X}_t^{\mathrm{sort}}\).
This relative ordering information is based purely on rank, independent of specific fitness scales, which may contribute to better cross-task generalization.

\textbf{Rank-fitness dual-aware attention.}
In standard self-attention, the input sequence $\mathbf{X}^{\mathrm{sort}}_t$ is linearly projected to form queries, keys, and values so as to model dependencies across different positions in the sequence.
Such learned projections can improve fitting to a specific problem but may hurt generalization when transferring to new tasks.
Therefore, we remove the three projections on the input sequence and compute attention directly from the sorted population representation.
Moreover, relying on ranking alone is a coarse representation: it reflects the relative ordering of individuals in the population, yet fails to characterize their fitness relations.
To this end, we further introduce fitness information to construct the attention-weight matrix.

Specifically, we first normalize the sorted fitness vector across individuals by
\begin{equation}
\tilde{\mathbf{F}}_t=\mathrm{softmax}(\mathbf{F}_t^{\mathrm{sort}}),
\end{equation}
so that $\tilde{\mathbf{F}}_t$ lies on the probability simplex. We then map $\tilde{\mathbf{F}}_t$ into fitness-aware queries and keys through two lightweight linear projections, and use the resulting attention weights to aggregate the sorted population:
\begin{equation}
\mathbf{X}_t^{\mathrm{fit}}
=
\mathrm{softmax}\!\left(
\frac{
(\tilde{\mathbf{F}}_t \mathbf{W}_f^{Q})
(\tilde{\mathbf{F}}_t \mathbf{W}_f^{K})^{\top}
}{
\sqrt{d_f}
}
\right)\mathbf{X}_t^{\mathrm{sort}}
\in \mathbb{R}^{N\times D},
\end{equation}
where $\mathbf{W}_f^{Q}, \mathbf{W}_f^{K}\in\mathbb{R}^{1\times d_f}$, $d_f$ is the fitness embedding dimension, and $\mathrm{softmax}(\cdot)$ is applied row-wise. In this way, the attention weights are determined solely by fitness, while the aggregated content remains in the original decision space. This design avoids task-specific projections on the value branch, reduces sensitivity to coordinate scaling, and improves cross-task generalization.

Furthermore, we introduce a trainable rank-based attention matrix $\mathbf{A}_{\mathrm{rank}} \in \mathbb{R}^{N \times N}$, which is learned purely from ordering information. We combine the rank-based and fitness-based branches as
\begin{equation}
\mathbf{X}_t^{c}
=
\lambda_{\mathrm{rank}}\,\mathbf{A}_{\mathrm{rank}}\mathbf{X}_t^{\mathrm{sort}}
+
\lambda_{\mathrm{fit}}\,\mathbf{X}_t^{\mathrm{fit}},
\end{equation}
where $\lambda_{\mathrm{rank}}$ and $\lambda_{\mathrm{fit}}$ are learnable fusion weights satisfying $\lambda_{\mathrm{rank}}+\lambda_{\mathrm{fit}}=1$.

\textbf{FFN perturbation with residual fusion.}
The attention layer aggregates information across the population via learned weights. However, its ability to explore new regions can be limited, which may lead to premature convergence on complex landscapes. To enhance exploration, we introduce a feed-forward network (FFN) to produce a learnable nonlinear perturbation of $\mathbf{X}_t^{c}$:
\begin{equation}
\mathbf{X}_t^{m}=\mathrm{FFN}(\mathbf{X}_t^{c}),
\end{equation}
where $\mathbf{X}_t^{m}$ denotes the population representation after applying the FFN perturbation to $\mathbf{X}_t^{c}$. This perturbation module encourages exploration by introducing nonlinear variations into the population features.

To preserve the original decision vectors, the dual-attention representation $\mathbf{X}_t^{c}$, and the nonlinear perturbation introduced by the FFN, we combine the three streams through a residual fusion scheme to form the refined population representation:
\begin{equation}
\mathbf{X}_{t}^{\mathrm{ref}}
=
\boldsymbol{\beta}_{1}\odot \mathbf{X}_{t}^{\mathrm{sort}}
+
\boldsymbol{\beta}_{2}\odot \mathbf{X}_{t}^{c}
+
\boldsymbol{\beta}_{3}\odot \mathbf{X}_{t}^{m},
\end{equation}
where $\odot$ denotes element-wise multiplication, and $\boldsymbol{\beta}_{1},\boldsymbol{\beta}_{2},\boldsymbol{\beta}_{3}\in\mathbb{R}^{N\times 1}$ are learnable fusion coefficients. This design enables the model to balance the original population information, the attention-based interaction features, and the nonlinear perturbation when constructing the refined population representation.

\subsection{Elite Synthesis Decoder}

Evaluating a large number of offspring as in conventional evolutionary algorithms would quickly exhaust the evaluation budget~\cite{Wang25}. We therefore introduce ESD as the decision module of NeurGO. It maps the refined population representation produced by PCE to a small set of high-quality candidates, so that only \(k\) candidate solutions synthesized from \(\mathbf{X}_t^{\mathrm{ref}} \in \mathbb{R}^{N \times D}\) are evaluated at each iteration.

\textbf{Multi-source feature concatenation.}
Although PCE produces a refined population representation $\mathbf{X}_t^{\mathrm{ref}}$ from population observations, it mainly characterizes the evolutionary tendency and the current search state. To provide richer references for elite generation, we additionally use the sorted raw population $\mathbf{X}^{\mathrm{sort}}_t$ and its normalized fitness $\tilde{\mathbf{F}}_t$ as priors, together with $\mathbf{X}_t^{\mathrm{ref}}$, for subsequent synthesis. These three sources are concatenated into an individual-wise feature matrix $\mathbf{Z}_t \in \mathbb{R}^{N \times (2D+1)}$, where $\mathbf{X}^{\mathrm{sort}}_t$ provides the population's decision-space locations, $\tilde{\mathbf{F}}_t$ provides relative fitness information among individuals, and $\mathbf{X}_t^{\mathrm{ref}}$ provides a refined population representation of the current search state.

\textbf{Elite candidate synthesis.}
ESD is designed to directly synthesize $k$ high-value candidates from the current population. We use an MLP $g_{\phi}(\cdot)$ to produce a score matrix that quantifies each individual's contribution to each synthesized candidate:
\begin{equation}
\mathbf{S}_t = g_{\phi}(\mathbf{Z}_t) \in \mathbb{R}^{N \times k},
\end{equation}
where $(\mathbf{S}_t)_{i,j}$ indicates the contribution score of the $i$-th individual to the $j$-th candidate. We then apply a softmax over the population dimension to normalize $\mathbf{S}_t$ and obtain the contribution weight matrix $\mathbf{P}_t \in \mathbb{R}^{N \times k}$, where each column $\mathbf{P}_t[:,j]$ forms a distribution over individuals indicating their relative contributions to the $j$-th candidate.

Given $\mathbf{P}_t$, we synthesize $k$ elite candidates by aggregating the refined population representation with the contribution weights:
\begin{equation}
\mathbf{X}_t^{\mathrm{cand}} = \mathbf{P}_t^{\top}\mathbf{X}_t^{\mathrm{ref}},
\end{equation}
where $\mathbf{X}_t^{\mathrm{cand}} \in \mathbb{R}^{k \times D}$ denotes the matrix of $k$ synthesized candidate solutions. Since each candidate is synthesized as a weighted aggregation of the current population representations, the decoder is naturally biased toward recombining information already supported by the population. This design keeps candidate generation anchored to the current population, helping avoid out-of-distribution candidates and reducing wasted evaluations in expensive optimization.

\subsection{Quality--Diversity Loss}
During meta-training, we optimize a QD objective that jointly enforces candidate quality and diversity. 
This discourages overly similar candidates, reduces redundant evaluations, and improves sample efficiency.

\textbf{Fitness normalization.}
Fitness scales may vary substantially across black-box tasks. If the loss is computed directly from raw fitness values, a few tasks with larger numerical scales may dominate training, weakening cross-task generalization. Therefore, we apply z-score normalization to fitness values within each independent optimization run to remove inter-task scale differences.

Given the population \(\mathbf{X} \in \mathbb{R}^{N \times D}\) at iteration \(t\), and the evaluated candidate set \(\mathbf{X}^{\mathrm{cand}} \in \mathbb{R}^{k \times D}\) for the current task, we denote their fitness vectors by \(\mathbf{f}^{\mathrm{pop}} = f(\mathbf{X}) \) and \(\mathbf{f}^{\mathrm{cand}} = f(\mathbf{X}^{\mathrm{cand}})\), respectively. We further standardize each fitness vector using z-score normalization.
Under a limited evaluation budget, if only \(r=\min\!\bigl(k,\, B-\mathrm{NFE}\bigr)\) candidates can be evaluated at the current iteration, we evaluate only the first \(r\) synthesized candidates, where \(\mathrm{NFE}\) denotes the number of expensive function evaluations consumed so far.
For clarity, we use \(k\) to denote the number of candidate solutions synthesized at each iteration.

\textbf{Best-value improvement loss.}
In expensive optimization, the primary objective is typically not to improve the average performance of all candidates, but to obtain a better incumbent solution with as few evaluations as possible. Accordingly, we define the quality term as the best-value improvement of the current candidate set over the parent population:
\begin{equation}
\mathcal{L}_{\mathrm{qual}}
=
\frac{1}{|\Omega|}
\sum_{\tau \in \Omega}
\left(
\min_{1\le j \le k}\tilde{\mathbf{f}}^{\mathrm{cand}(\tau)}_j
-
\min_{1\le l \le N}\tilde{\mathbf{f}}^{\mathrm{pop}(\tau)}_l
\right),
\end{equation}
where $\Omega$ denotes the mini-batch of meta-training tasks and $|\Omega|$ is its size. For each task $\tau\in\Omega$, $\tilde{\mathbf{f}}^{\mathrm{cand}(\tau)}$ and $\tilde{\mathbf{f}}^{\mathrm{pop}(\tau)}$ are the z-score normalized fitness vectors of the evaluated candidates and the parent population, respectively. Thus, $\mathcal{L}_{\mathrm{qual}}$ measures the average best-value improvement achieved by the candidate set over the current population across tasks.

\textbf{Candidate-set diversity loss.}
In generative optimization, optimizing only the quality term may cause the model to produce highly homogeneous candidates. Although such candidates may occasionally improve the best value, they also reduce diversity and may waste evaluations through redundancy. To mitigate this issue, we introduce a diversity regularization term that encourages the candidates to spread out in the decision space. For each task $\tau\in\Omega$, let $\mathbf{X}^{\mathrm{cand}(\tau)}\in\mathbb{R}^{k\times D}$ denote the evaluated candidate set. We define the diversity loss as the negative mean pairwise Euclidean distance among candidates, normalized by the domain scale:
\begin{equation}
\mathcal{L}_{\mathrm{div}}
=
-\frac{1}{|\Omega|}
\sum_{\tau\in\Omega}
\frac{1}{k^{2}}
\sum_{i=1}^{k}\sum_{j=1}^{k}
\frac{\left\|\mathbf{x}^{\mathrm{cand}(\tau)}_{i}-\mathbf{x}^{\mathrm{cand}(\tau)}_{j}\right\|_{2}}
{(x_{\mathrm{ub}}-x_{\mathrm{lb}})\sqrt{D}+\epsilon},
\end{equation}
where $\|\cdot\|_2$ denotes the $\ell_2$ norm, $(x_{\mathrm{lb}},x_{\mathrm{ub}})$ are the box bounds of the search space, and $\epsilon>0$ is a small constant for numerical stability. The negative sign encourages the candidates to maintain a larger average pairwise distance in the decision space.

\textbf{Overall objective.}
We combine the two terms into the final training objective, $\mathcal{L}_{\mathrm{QD}}=\mathcal{L}_{\mathrm{qual}}+\lambda\,\mathcal{L}_{\mathrm{div}}$, where $\lambda \ge 0$ controls the trade-off between quality and diversity. We linearly decay $\lambda$ during training to encourage broader exploration in the early stage and gradually shift the emphasis toward convergence.
The pseudocode of the training procedure is provided in the supplementary material available in our Zenodo repository.

\section{Experiments}
\subsection{Experimental Setup}
\textbf{Datasets.} 
We follow B2Opt to build a surrogate-function training set with three functions. 
Training optimizes the surrogate objective, and transfer is assessed on the true objectives from two held-out benchmarks: CEC 2008 ($F_1$--$F_6$)~\cite{Tang07} and COCO BBOB ($f_1$--$f_{24}$)~\cite{Hansen21}.
During training, we optimize computationally inexpensive surrogate objectives to reduce evaluations of the true objective, and at test time we assess transfer on the original benchmark functions. 
We fix the dimensionality to $D=10$ in all experiments. The three training surrogate functions are defined on $[-10,10]^D$. For testing, BBOB uses the standard search domain $[-5,5]^D$, while the CEC functions follow their original function-dependent domains. 
To match the low-budget expensive-optimization setting, we set the maximum number of true function evaluations to $\mathrm{MaxFE}=300$ for all experiments.
Additional details of the CEC and COCO BBOB test functions are provided in the supplementary material available in our Zenodo repository.

\textbf{Baselines.} To obtain comprehensive and representative conclusions, we compare NeurGO against a broad set of baselines covering four classical paradigms. 
\textbf{EAs:}
We include classical population-based optimizers without surrogate modeling, including DE~\cite{Storn97}, GA, SHADE~\cite{Tanabe13}, and CMA-ES~\cite{Hansen01}.
\textbf{SAEAs:}
We consider EGO~\cite{Jones98} and SACOSO~\cite{Sun17}, which improve sampling efficiency under limited evaluation budgets by learning a cheap approximation of the objective and using it to guide candidate selection.
\textbf{BO:}
We compare against TurBO~\cite{Eriksson19} and HEBO~\cite{CowenRivers22}, two strong BO variants designed for challenging settings (e.g., high dimensionality and noisy observations) through localized modeling and robust acquisition optimization.
\textbf{MetaBBO:}
We further include LGA~\cite{Lange23}, LES~\cite{Lange23a}, and B2Opt as representative learning-based optimizers that learn search strategies from prior tasks and transfer them to new optimization problems.
All baselines are evaluated under the same evaluation budget and independently run 10 times to ensure a fair comparison. 
Detailed baseline configurations are provided in the supplementary material available in our Zenodo repository.

\subsection{Main Results}
Due to the limited space, additional experimental results, including detailed statistics, high-dimensional results, and supplementary analyses that are not included in this paper are provided in our Zenodo repository at \url{https://doi.org/10.5281/zenodo.19644601}.

\textbf{NeurGO vs.\ EAs.}
Compared with four classical evolutionary algorithms (EAs), Table \ref{table1} shows that NeurGO delivers superior performance on the majority of test functions, while underperforming on a small subset of particularly difficult cases. 
This observation is further supported by higher win counts against EAs and a better average rank across the full test suite.
We attribute NeurGO's advantage over EAs primarily to the difference in search mechanisms. 
Conventional EAs explore a large candidate pool through variation operators and selection.
Under tight budgets, many evaluations may be spent on low-quality individuals.
In contrast, NeurGO leverages the learned population-state representation to directly generate a small set of high-value candidates, which improves evaluation efficiency. 

\textbf{NeurGO vs.\ SAEAs.}
Among surrogate-assisted evolutionary baselines, Table~\ref{table1} indicates that NeurGO achieves better performance and exhibits greater stability on most test functions, whereas EGO or SACOSO is slightly superior on a small subset. 
A key reason is the strict evaluation budget. EGO and SACOSO repeatedly refit surrogates from very few true evaluations, which can accumulate model errors and misguide sampling decisions, increasing performance variability.
By comparison, NeurGO avoids online surrogate fitting. It generates a small set of high-quality candidates from population-state representations learned across tasks and maintains more consistent progress under limited budgets.

\begin{table}[t]
\centering
\caption{Overall performance comparison on BBOB and CEC benchmarks. The w/l column indicates the number of wins and losses against NeurGO, and Avg. Rank denotes the average ranking across all test functions.}
\label{table1}
\small
\setlength{\tabcolsep}{6pt}
\renewcommand{\arraystretch}{1.15}
\begin{tabular}{lccc}
\toprule
Algorithm & BBOB (w/l, rank) & CEC (w/l, rank) & Avg. Rank \\
\midrule
DE     & 5/19 (4.8)  & 0/6 (10.2) & 7.5 \\
GA     & 4/20 (8.2)  & 0/6 (8.0)  & 8.1 \\
SHADE  & 4/20 (8.6)  & 0/6 (8.8)  & 8.7 \\
CMA-ES  & 2/22 (10.1) & 0/6 (11.3) & 10.7 \\
SACOSO & 4/20 (6.6)  & 0/6 (5.8)  & 6.2 \\
EGO    & 6/18 (4.8)  & 0/6 (3.8)  & 4.3 \\
HEBO   & 7/17 (2.5)  & 0/6 (2.3)  & 2.4 \\
TurBO  & 6/18 (3.2)  & 0/6 (4.3)  & 3.8 \\
B2Opt  & 5/19 (5.7)  & 0/6 (3.7)  & 4.7 \\
LGA    & 4/20 (10.0) & 0/6 (7.7)  & 8.9 \\
LES    & 3/21 (10.2) & 0/6 (11.0) & 10.6 \\
\midrule
NeurGO & --/-- & --/-- & \textbf{2.0} \\
\bottomrule
\end{tabular}
\end{table}

\textbf{NeurGO vs.\ BO.}
As shown in Table \ref{table1}, NeurGO is overall superior to both Bayesian optimization baselines. 
NeurGO consistently surpasses TurBO and HEBO on the CEC suite and achieves better final performance on most BBOB functions.
However, TurBO and HEBO remain competitive on a small subset of BBOB functions with more complex global structure and many misleading local minima.
In general, NeurGO demonstrates a more stable advantage, whereas BO performance is more sensitive to surrogate accuracy and acquisition quality under limited observations.
We attribute this difference to how the two paradigms utilize information under tight budgets. With a small evaluation budget,  acquisition optimization can be noisy and multimodal.
TurBO’s trust-region updates and HEBO’s acquisition optimization can further bias evaluations toward limited regions.
In contrast, NeurGO conditions on population states and generates a small set of candidates per step, which better preserves exploration and provides more stable improvements.

\textbf{NeurGO vs.\ MetaBBO.}
As summarized in Table \ref{table1}, NeurGO achieves better final performance on the vast majority of test functions and exhibits notably stable results on the CEC meta-test functions (F1--F6), suggesting stronger cross-function generalization. 
B2Opt is designed for low-budget optimization by learning search operators, whereas LGA and LES use meta-learning to discover or modify update rules for evolutionary algorithms.
Instead, NeurGO directly synthesizes a small set of high-value candidates from population states, enabling more effective use of each evaluation under strict budgets.
We also observe that MetaBBO methods remain competitive on a small subset of functions with more complex global structure and many local optima.
Nonetheless, NeurGO is more robust and delivers more reliable performance in the low-budget regime.

\begin{figure}[h]
    \centering
    \includegraphics[width=0.70\linewidth]
    {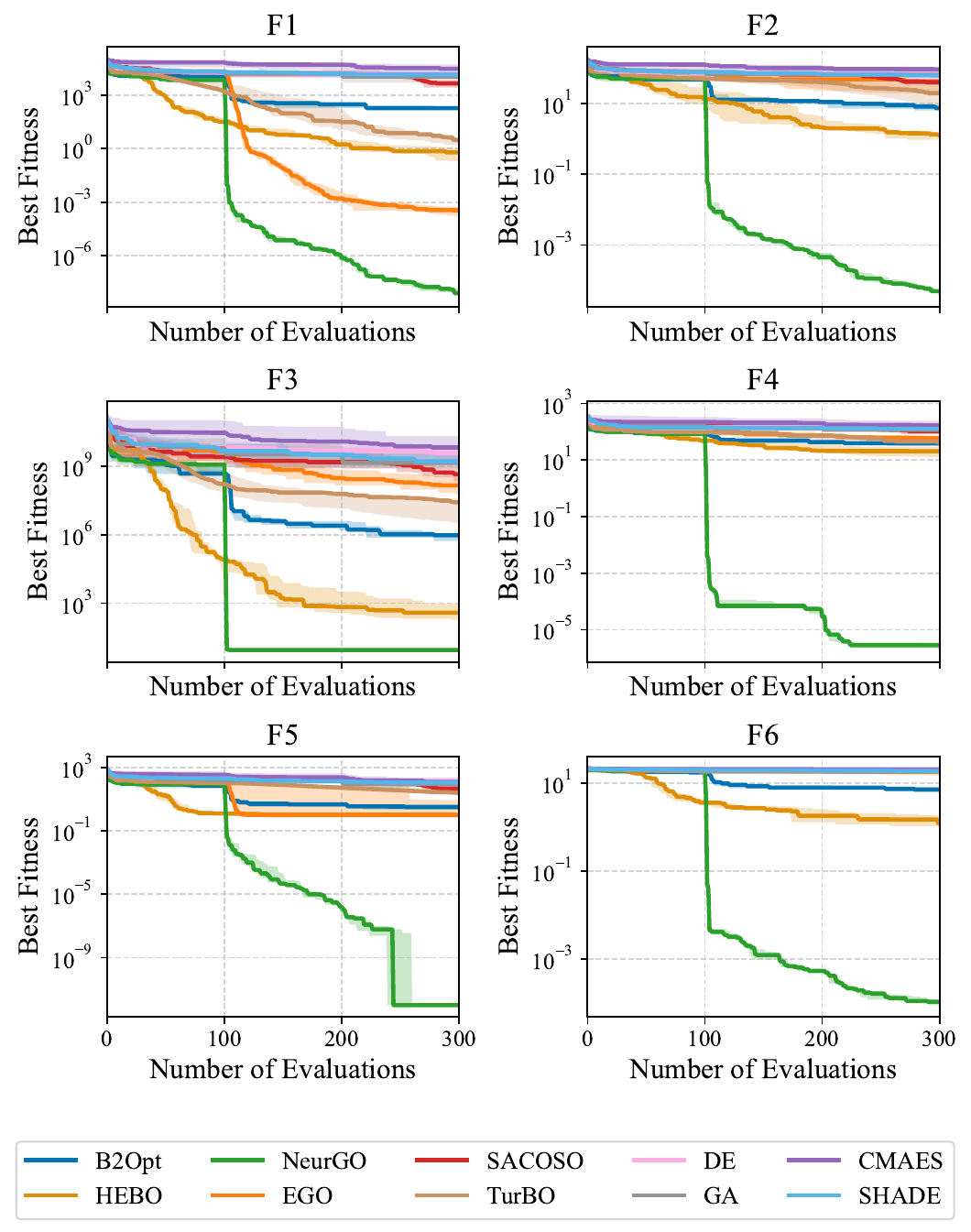}
    \caption{Convergence performance of NeurGO and other baselines on the CEC functions.}
    \label{fig4:convergence_cec}
\end{figure}

\textbf{Convergence analysis.}
Figure \ref{fig4:convergence_cec} shows that NeurGO converges faster than all baselines across the six CEC functions. 
It exhibits a sharp fitness drop around 100 evaluations and then continues improving, while most methods struggle to achieve effective convergence under the same budget.
This observation suggests that NeurGO uses scarce evaluations more efficiently by turning them into higher-quality candidate proposals.
A plausible explanation is that meta-training induces a learned prior over promising search regions, which helps NeurGO localize good solutions early and keep refining them throughout the search.

\begin{figure}[t]
    \centering
    \includegraphics[width=0.65\linewidth]
    {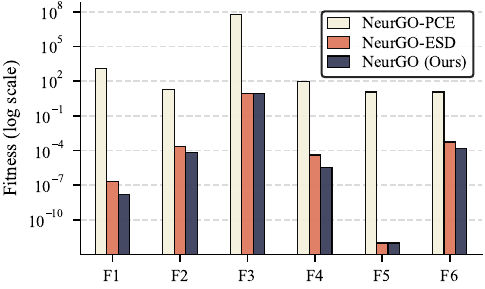}
    \caption{Ablation study of NeurGO components on CEC functions. Fitness values are shown on a log scale (lower is better).}
    \label{fig3:ablation}
\end{figure}

\subsection{Ablation Study}

\textbf{Effectiveness of PCE.}
To assess the contribution of PCE, we remove its representation enhancement and pass the initial population directly to the candidate-generation stage.
We further remove the feature vector generated by PCE, which is used as the base vectors in ESD’s weighted aggregation.
Consequently, ESD computes aggregation weights using only the sorted population vectors and their fitness values, and generates $k$ candidates by weighted recombination over the current population.
Figure \ref{fig3:ablation} shows that this ablation degrades performance substantially on almost all test functions, indicating that the population context representation learned by PCE is crucial for generating high-quality candidates.
It prevents ESD from merely performing a simple recombination of the current population, and instead enables more directed candidate synthesis conditioned on the search state.
We additionally ablate the attention module, FFN, and residual connections within PCE, and observe consistent performance drops when removing any component. 
Detailed ablation results are provided in the supplementary material available in our Zenodo repository.

\textbf{Effectiveness of ESD.}
The representation produced by PCE can be regarded as a set of latent candidate solutions.
To evaluate the contribution of ESD, we remove the MLP in ESD that learns the aggregation weights and instead randomly sample $k$ candidates from this set for evaluation.
This variant performs worse on most functions, indicating that random sampling rarely selects consistently high-quality candidates under limited budgets. 
ESD alleviates this problem by learning aggregation weights that allocate evaluations to more promising candidates and stabilize optimization progress.
These results highlight that ESD’s learned aggregation is essential for effectively exploiting PCE’s latent space, rather than relying on unguided selection.

\begin{figure}[h]
    \centering
    \includegraphics[width=0.65\linewidth]
    {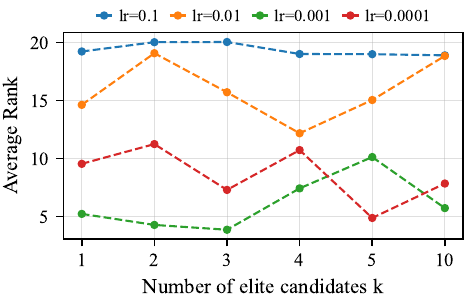}
    \caption{Hyperparameter sensitivity of NeurGO to the learning rate and the number of elite candidates, measured by the average rank across test functions.}
    \label{fig5:hyper_params}
\end{figure}
\subsection{Hyperparameter Sensitivity}

\textbf{Impact of learning rate.}
We vary the learning rate and assess its effect on final performance.
Figure~\ref{fig5:hyper_params} reports the average rank across test functions for different combinations of the learning rate \(lr\) and the number of elite candidates \(k\), where a lower rank indicates better overall performance.
The results show that \(lr=0.001\) consistently achieves the lowest average rank across different choices of \(k\), while both larger and smaller rates result in inferior performance.

\textbf{Impact of candidate number.}
We vary $k$ while keeping the total evaluation budget fixed to study how the number of candidate solutions generated per iteration affects exploration of the search space and the allocation of evaluations.
The results show that small $k$ explores fewer regions of the search space and may miss promising areas.
Conversely, a very large $k$ may allocate fewer evaluations to each candidate on average, which reduces the average candidate quality.
These findings suggest that a proper trade-off between diversity and candidate quality is essential.
In our experiments, $k=3$ provides competitive performance across test functions.

\section{Conclusion}
In this paper, we propose NeurGO, a generative learning-based MetaBBO framework for expensive black-box optimization. 
NeurGO directly synthesizes a limited set of elite candidates from historical population states, alleviating the need for repeated online surrogate fitting and acquisition function optimization under a low evaluation budget. 
By modeling population context and generating candidates in an end-to-end manner, NeurGO concentrates the limited evaluation budget on promising regions while maintaining adequate exploration. 
Extensive experiments on the CEC and COCO BBOB benchmarks demonstrate that NeurGO achieves superior performance against four classes of baselines under limited evaluation budgets, while additional high-dimensional results provided in the supplementary material available in our Zenodo repository offer further evidence of its scalability.
In future work, we plan to extend NeurGO to broader task distributions and more diverse training suites, and to further investigate its applicability in more challenging expensive optimization settings. We hope NeurGO promotes the integration of generative methods into MetaBBO and facilitates more effective approaches to expensive black-box optimization.

\begin{credits}
\subsubsection{\ackname} 
This work was partly supported by the National Natural Science Foundation of China under Grant Nos. 62403442 and 62576325, the General Program of Hubei Provincial Natural Science Foundation (JCZRMS202600202), the Postdoctoral Fellowship Program of CPSF under Grant No. GZC20241600, the "CUG Scholar" Scientific Research Funds at China University of Geosciences (Wuhan) (Project No. 2024011), as well as support from the Hubei Province Postdoctoral Talent Introduction Program (Project No. 2024HBBHJD096).
\end{credits}
%
%
%
\bibliographystyle{splncs04}
\bibliography{exsamplepaper}
%




\end{document}